\documentclass{bmvc2k}
\usepackage{times}
\usepackage{epsfig}
\usepackage{graphicx}
\usepackage{algorithm}
\usepackage[noend]{algpseudocode}
\usepackage{amsmath}
\usepackage{amssymb}
\usepackage{amssymb}
\usepackage{xcolor}
\usepackage{pifont}
\newcommand{\cmark}{\ding{51}}%

\definecolor{captionblue}{rgb}{0,0,0.4}
\usepackage[font={color=captionblue,small},skip=1pt]{caption}

\title{MagnifierNet: Towards Semantic Adversary and Fusion for Person Re-identification}

\addauthor{Yushi Lan*, Yuan Liu*}{lanyushi15@gmail.com, yliu050@e.ntu.edu.sg}{1}
\addauthor{Xinchi Zhou, Maoqing Tian, Xuesen Zhang, Shuai Yi}{zhouxinchi, tianmaoqing, zhangxuesen, yishuai@sensetime.com}{1}
\addauthor{Hongsheng Li}{hsli@ee.cuhk.edu.hk}{2}

\addinstitution{
SenseTime Group Limited
}
\addinstitution{
The Chinese University of HongKong
}

\runninghead{LAN,LIU ETAL}{MagnifierNet}

\begin{document}

\maketitle

\begin{abstract}
Although person re-identification (ReID) has achieved significant improvement recently by enforcing part alignment, it is still a challenging task when it comes to distinguishing visually similar identities or identifying the occluded person. In these scenarios, magnifying details in each part features and selectively fusing them may provide a feasible solution. In this work, we propose MagnifierNet, a triple-branch network which accurately mines details from whole to parts. Firstly, the holistic salient features are encoded by a global branch. Secondly, to enhance detailed representation for each semantic region, the ``Semantic Adversarial Branch" is designed to learn from dynamically generated semantic-occluded samples during training. Meanwhile, we introduce ``Semantic Fusion Branch" to filter out irrelevant noises by selectively fusing semantic region information sequentially. To further improve feature diversity, we introduce a novel loss function ``Semantic Diversity Loss" to remove redundant overlaps across learned semantic representations. 
State-of-the-art performance has been achieved on three benchmarks by large margins. Specifically, the mAP score is improved by 6\% and 5\% on the most challenging CUHK03-L and CUHK03-D benchmarks.
\end{abstract}

\section{Introduction}
\label{sec:intro}
Person Re-IDentification (ReID) has attracted significant attention for its critical role in video surveillance and public security. Specifically, given a query image, the ReID system amounts to retrieve all the images of the same identity according to their semantic similarity from a large gallery, typically captured by distinctively different cameras from various viewpoints.


Although advancement has been witnessed, 
there are still some challenging issues to be better solved, such as pose variation, body occlusion and the part misalignment. Former works attempt to tackle these problems via methods including attention mechanism~\cite{li2018harmonious,zhao2017deeply}, body feature cropping~\cite{Zhao_2017_CVPR,Sun_2018_ECCV}, and human parsing~\cite{Kalayeh_2018_CVPR}. Most of these works mainly focus on the enhancement of feature alignment, while the exploitation of fine-grained details is often ignored. Detailed semantic information, however, is critical for further improvements, especially in cases where identities are heavily occluded or only share minor differences in certain parts. 

To effectively address the challenges mentioned above, we propose MagnifierNet, a novel triple-branch network that not only extracts aligned representation but also magnifies fine-grained details and selectively fuses each semantic region. The framework is constructed from whole to parts.  Apart from holistic representation provided by conventional global branch, the Semantic Adversarial Branch ($SAB$) learns fine-grained representation for each semantic region, and the Semantic Fusion Branch ($SFB$) fuses each semantic feature selectively to focus only on beneficial information. Meanwhile, a light-weight mask module is applied to impose alignment constraint on the feature map and provide segmentation masks during inference. In this way, our model is able to distinguish similar identities, recognize occluded samples with semantic details, as well as alleviate negative influence from noises.

Notably, the semantic features under different body part masks might share overlapping information, as these masks are generated at high-level feature maps with large receptive fields. These overlaps could reduce the network’s capability to capture distinctive details of different semantic regions. Hence, we propose a novel Semantic Diversity ($SD$) Loss to improve diversity among different parts, which further enhances the network performance as shown in our experiments. 

Our main contribution can be summarized as follows: 
\begin{itemize}
\item We propose a novel Semantic Adversarial Branch that magnifies fine-grained details in each semantic region and regularizes the learning of ReID.
\item We design a novel Semantic Fusion Branch which selectively fuses semantic information sequentially, focusing on beneficial features while filtering out noises.
\item We further improve semantic feature diversity by introducing a novel Semantic Diversity Loss.
\end{itemize}
Experimental results show that our proposed MagnifierNet achieves state-of-the-art performance on three ReID benchmark datasets including Market-1501, DukeMTMC-reID, and CUHK03-NP. 

\section{Related Work}
\noindent\textbf{ReID and Part Based Methods}   Person ReID task aims to retrieve target images belonging to the same person based on their similarity. In terms of \textit{feature representation}, the great success of deep convolution networks has pushed ReID benchmarks to a new level~\cite{Sun_2018_ECCV, zheng2018discriminatively, chen2018group, zhao2017deeply}. Recently, researchers have intensively focused on extracting local person features to enforce part alignment and feature representation capability. For instance, Zhao \emph{et al.}~\cite{Zhao_2017_CVPR} proposed a body region proposal network which utilized human landmark information to compute relative aligned part representation. Zhao \emph{et al.}~\cite{zhao2017deeply} shared a similar idea, but with feature representation generated from K part detectors. Sun \emph{et al.}~\cite{Sun_2018_ECCV} proposed a Part-based Convolution Baseline (PCB) network which focuses on the consistency of uniform partition part-level feature with refined stripe pooling. Based on PCB, MGN~\cite{wang2018learning} and Pyramidal Model~\cite{zheng2018pyramidal} explored multi-branch networks to learn features of different granularity, which attempted to incorporate global and local information. However, their methods result in complicated structures and seriously increased parameters, and also lacks the ability to represent accurate human regions. In comparison, our approach aggregates multi-granularity features in a more efficient way, resulting in smaller network size and better performance.

\noindent\textbf{Semantic Aligned ReID}   Considering serious occlusion under practical surveillance scenarios and the highly structured composition of the human body, some researchers aim to push the margin of ReID task via guided feature alignment. The development of human parsing methods~\cite{Gong_2017_CVPR,Kalayeh_2018_CVPR} and pose estimation~\cite{Xiao_2018_ECCV, Cao_2017_CVPR, Xu_2018_CVPR} facilitate the use of human semantic information as external cues to promote the performance of ReID task. Xu \emph{et al.}~\cite{Xu_2018_CVPR} resort to the assistance of predicted keypoints confidence map to extract aligned human parts representation. Similarly, Huang \emph{et al.}~\cite{huang2018eanet} proposed part-aligned pooling based on delimited regions which shows significant improvement in cross-domain ReID task. Kalayel \emph{et al.}~\cite{Kalayeh_2018_CVPR} adopts predefined body regions as supervision to drive the model to learn alignment representation automatically. However, these methods treat each semantic region equally and did not further explore the details in each region.

\noindent\textbf{Metric Learning for ReID}   With the adoption of large datasets and deeper convolution networks, ReID task has gradually evolved to a metric learning problem, which aims to retrieve target images based on their feature similarity. For this purpose, recent works in metric learning have paid intensive attention to loss function design, such as triplet loss~\cite{DING20152993}, center loss~\cite{wen2016discriminative} and margin-based loss~\cite{Liu_2017_CVPR, Wang_2018_CVPR, Deng_2019_CVPR}, etc. Hard triplet mining strategy has also been successfully utilized in ReID~\cite{cheng2016person,dai2019batch}. Nevertheless, the above-mentioned approaches mostly focus on image-level feature refinement. Meanwhile, orthogonality constraints are explored by~\cite{sun2017svdnet} for ReID to encourage the learning of informative and diverse features. To further exploit the motivation, the Semantic Diversity Loss in our work encourages orthogonality directly on semantic feature level, which accurately improves diversity among body regions that are of great significance for the ReID task.
\section{Methodology}

\begin{figure*}[t!]
        \center{\includegraphics[width=1.0\textwidth]
        {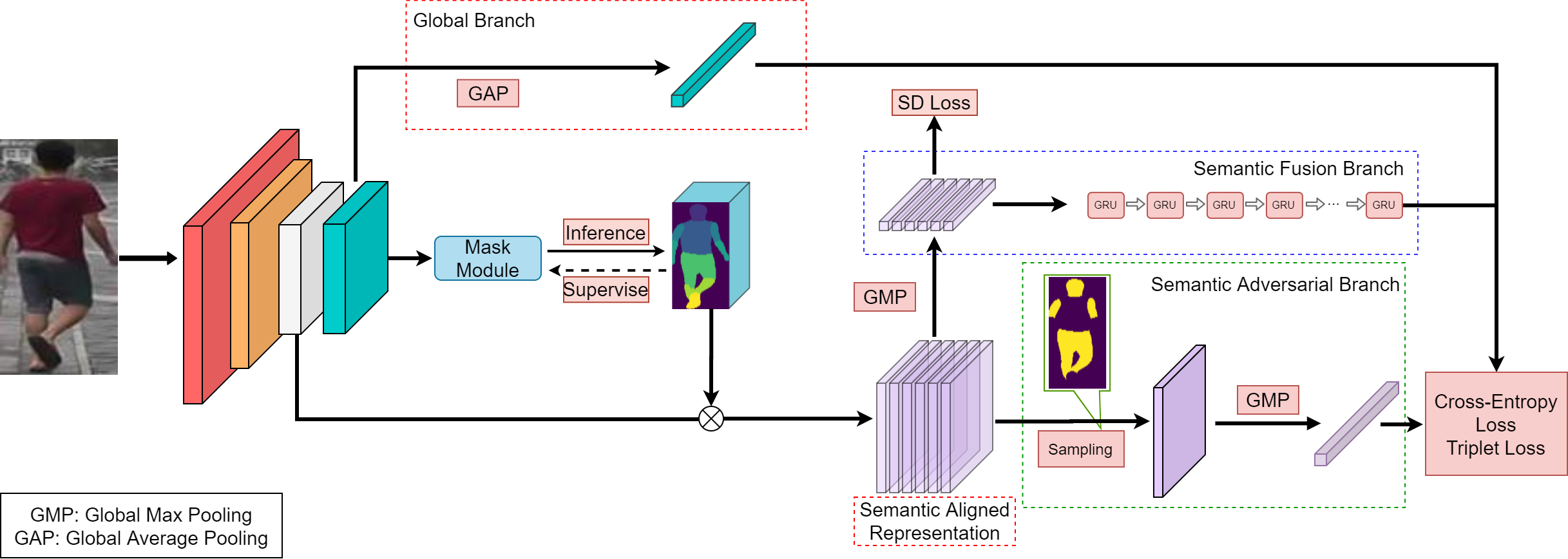}}
        \caption{\label{struct}The overall architecture of the proposed MagnifierNet. The Semantic Adversarial Branch explores the hidden details in each semantic region by limiting available information for training. The Semantic Fusion Branch encodes beneficial features from each semantic region sequentially while filtering out noises. The $SD$ Loss further improves representation diversity among all semantic features.}
\end{figure*}

This section introduces our proposed framework, including the Semantic Adversarial Branch ($SAB$) to magnify attentive feature of each region, the Semantic Fusion Branch ($SFB$) to fuse and filter extracted features and a commonly adopted Global Branch to capture holistic representations. A light-weight mask module ($M$) is introduced to parse semantic information of human parts for $SAB$ and $SFB$. The whole network is trained end-to-end in the first training stage. The Semantic Diversity Loss is then added to fine-tune the network in the second training stage. The overview of the proposed network structure is shown in Figure~\ref{struct}.

\subsection{Mask Module and Semantic Aligned Representation} 
The main idea of the mask module is to extract accurate human landmark features, through which pixel-level representations belonging to the same human semantic landmark can be aggregated. Given the binary segmentation masks for all body parts $R_{seg}$ and the feature map $F_{cnn}$ of a given pedestrian image $x$, we can generate the \textit{Semantic Aligned Representation} via the outer product ($\bigotimes$). The semantic information can then be further leveraged by other components of the network, along with a pooling operation. We adopt global max-pooling operation ($GMP(.)$) in our experiments.

Specifically, we rescale $R_{seg}$ into the same size as $F_{cnn}$ using bilinear interpolation and calculate the \textit{Semantic Aligned Representation} $X^{ID}$ as shown in Equation~\ref{eq:1},
\begin{equation} \label{eq:1}
 X^{ID}=\{{x_{k}^{ID}}\}_{k=1}^{K}=F_{cnn} \bigotimes R_{seg}
\end{equation}
where $K$ is the region number and ${x_{k}^{ID}}$ is essentially the aligned representation of the semantic region it belongs to. $x_{k}\in{R^{c}}$ and $c$ is the channel number. For the pixels predicted as background, we still aggregate the representations of all pixels as the background can provide complementary information and augment the original representations.

We obtain $R_{seg}$ through a pre-trained Human Parsing model~\cite{fu2018dual} for all datasets as the pseudo mask label during training. To obtain $R_{seg}$ during inference and reinforce alignment across identities~\cite{huang2018eanet}, we include a light-weight segmentation head after backbone output in the mask module with the structure proposed in~\cite{he2017mask}. It predicts $K$ human region masks from input feature maps, and is supervised by the generated pseudo labels during training. More complex segmentation head designs have the potential to improve performance but are not the focus of our work.

\subsection{Semantic Adversarial Branch}
Though \textit{Semantic Aligned Representations} have been acquired, we argue that detailed feature representations haven't been captured in all semantic regions. Further magnification of semantic details is necessary to maintain robust model performance, especially in crowded venues where the identities are usually partially visible. 
Inspired by the success of adversarial samples, we propose \textbf{Semantic Adversarial Branch} ($SAB$) which dynamically generates semantic-level perturbed samples to drive the network to learn discriminative representation for each semantic region. These adversarial samples are generated online and trained along with original images in a separate branch, which provides complementary information to the network training. 

\noindent\textbf{Adversarial Sample Generation}   Given an input image, we dynamically occlude some semantic regions with a constant following certain strategy. The samples we generate are adversarial to the network, but with more analogy to real-world scenarios which can drive the network to learn more attentive features of each semantic region. To exploit the advantages of the multi-branch network and utilize the computed \textit{Semantic Aligned Representations}, we directly generate adversarial samples on learned feature maps.

Concretely, given an image in a batch, we sample a semantic region group $\hat{O}$ made up of aligned semantic representations $X^{ID}$ formulated in Equation~\ref{eq:1}. Semantic aligned feature maps corresponding to the selected regions in $\hat{O}$ are zeroed out. This operation is applied to all feature channels and the generated adversarial samples of an identity can be formulated as Equation~\ref{eq:2},

\begin{equation}
F_{cnn}^{adversarial}= GMP(\sum_{k\notin{\hat{O}}} x_{k}^{ID})
\label{eq:2}
\end{equation}

\noindent\textbf{Region Sampling Strategy}   The intuition behind the occlusion sampling is to reinforce attentive representation learning of all regions via erasing the same semantic aligned features in each batch. Therefore, the most straightforward option is to randomly sample and occlude region(s) from $X^{ID}$ and only keep $\hat{K}$ regions. We call this sampling strategy \textbf{Random-baseline}.

Beyond this, we note that the remaining semantic regions should maintain sufficient yet dynamic cues to represent correct identity to drive the network to learn fine-grained representation for all semantic regions. Therefore, we manually split the aligned semantic regions into upper torso regions $U_{torso}$ 
and lower torso regions $L_{torso}$. 
In this way, we can perform sampling in two groups independently, avoiding the situations where the feature of an identity may lose discrimination under \textbf{Random-baseline} strategy. Specifically, we sample in a way such that equal number of regions remain in $U_{torso}$ and $L_{torso}$, namely $\hat{K}/2$ each where $\hat{K}$ is an even number. We name this sampling strategy \textbf{Random-torso}, which shows better results in our experiments.

As semantic regions are dynamically occluded in each batch,
limited number of semantic regions are left to represent an identity. Hence, the network is forced to capture fine-grained details in each semantic region instead of relying on certain discriminative region(s). Therefore, the model can learn more detailed and attentive representations on viable semantic regions and naturally alleviate occlusion or similar identity problems. 

\subsection{Semantic Fusion Branch}
To yield correct predictions even in cases of similar identities or noisy backgrounds, the model is supposed to focus more on informative semantic regions instead of equally considering noisy parts with misleading information. 

To address this problem, we propose the \textbf{Semantic Fusion Branch} ($SFB$) to selectively fuse all semantic regions, focusing on informative features while filtering out noises. Considering the gated units of GRU~\cite{cho2014properties} can estimate the importance of a given semantic feature conditioned by itself and other encoded features, it is natural to apply GRU to encode these semantic features sequentially to focus more on relevant information. 

We perform $GMP(.)$ on the \textit{Semantic Aligned Representation} to obtain a feature for each body part, and then feed the semantic aligned features sequentially into a one-layer GRU cell, and adopt the last GRU output as the semantic fusion of the input image. To compute the loss, we adopt the ``BNNeck" method proposed by~\cite{Luo_2019_CVPR_Workshops} to jointly minimize cross-entropy loss and triplet loss. Besides, we apply ReID supervision for each semantic feature to stabilize the training process, but the separate features are not utilized during inference.

 Notably, the two branches $SFB$ and $SAB$ share the same backbone features and segmentation masks. Hence, $SFB$ benefits directly from $SAB$'s semantic detail magnification process. As the feature representation under each semantic mask grows more informative, $SFB$ is able to accurately capture this information and selectively fuses them into a better feature embedding for the ReID task.

\subsection{Semantic Feature Diversification}
As the semantic masks are generated at high-level feature maps, features under different semantic masks might share overlapping information due to large receptive fields, which reduces the network's capability to capture details from different semantic regions. To improve diversity among semantic representations, we propose a novel \textbf{Semantic Diversity Loss} ($SD$ Loss) to remove redundancy among region features.

For a pair of semantic regions, we aim to diversify them by increasing their distance in the feature space. Specifically, we minimize the pair-wise cosine similarity among all semantic features as shown in Equation~\ref{sdloss}. 
\begin{equation}
L_{SD} = \frac{1}{N\tbinom{K}{2}} \sum_{k=1}^{N} \sum_{j=1}^{K} \sum_{i=j}^{K} {\frac{(p(x_{i}^{ID}))^Tp(x_{j}^{ID})}{max(\|{p(x_{i}^{ID})}\|\times\|{p(x_{j}^{ID})}\|, \epsilon)}}
\label{sdloss}
\end{equation}where \(x_{i}^{ID}\) and \(x_{j}^{ID}\) are a pair of semantic aligned features and $p$ stands for pooling operation implemented by $GMP(.)$. $N$ is batch size, $K$ is the number of semantic regions, \(T\) is matrix transpose, \(\|\|\) is vector magnitude and \(\epsilon\) is a small positive value used to avoid division by zero. 

As $SFB$ naturally generates individual semantic features $p(x_{k}^{ID})$ from \textit{Semantic Aligned Representation}, we calculate $SD$ Loss via $SFB$ on these pooled semantic features. Notably, as $SFB$ and $SAB$ share the same backbone layers and segmentation masks, minimizing $SD$ Loss through $SFB$ forces the backbone to extract diverse features which supports $SAB$ to further magnify semantic details. The overall loss function of the network is shown in Equation~\ref{overall-loss}, which is shared by all the network branches.

\begin{equation}
L_{Total} = L_{cls} + L_{tri} + \gamma L_{SD} + \lambda L_{mask},
\label{overall-loss}
\end{equation} where the first two terms stand for cross-entropy and triplet loss that optimize all branches, $L_{SD}$ is $SD$ Loss and $L_{mask}$ distills segmentation information which is used during inference. Weight coefficients $\gamma$ and $\lambda$ balance the importance of semantic diversification and segmentation respectively.



\begin{figure*}[t!]
        \center{\includegraphics[width=1.0\textwidth]
        {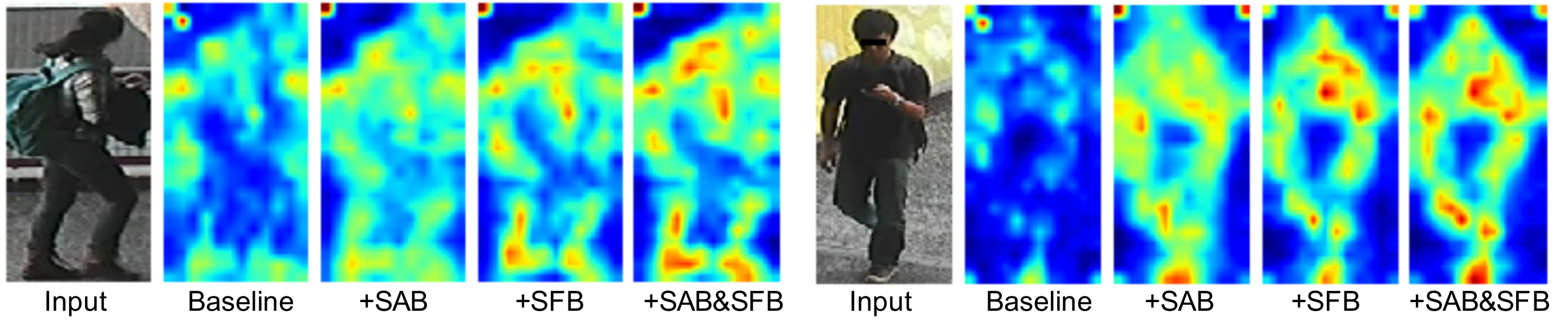}}
        \caption{\label{backbone-vis}Visualization of feature map activation with different network components. Both Semantic Adversarial and Fusion Branch are able to help the network capture fine-grained semantic details.}
\end{figure*}
\section{Experiments}
\subsection{Implementation Details}
We implement our model with reid-strong-baseline~\cite{Luo_2019_CVPR_Workshops} framework and train it on a single Tesla V100 GPU. Both $SAB$ and $SFB$ are connected to ResNet-50 stage 3's output with the final down-sampling layer removed to maintain more spatial information. All person images are re-sized to $384 \times 192$ as in~\cite{dai2019batch}. Coefficient $\lambda$ is set to $2$ and $\gamma$ is set to $2e-3$, which yields competitive results across all datasets. The training procedure is two-stage. Firstly, the model is trained near convergence without $SD$ Loss for 460 epochs. We simply adopt the optimizer settings of reid-strong-baseline~\cite{Luo_2019_CVPR_Workshops} for fair comparison. Then we fine-tuned the network on Equation~\ref{sdloss} for another 40 epochs with learning rate 0.001. The feature embeddings from all branches are concatenated as the final person representation during inference. The whole training procedure has 500 epochs and can be finished within 4 hours.
\begin{table*}[h!]
\begin{center}
\begin{tabular}{|c|c c|c c|c c|c c|}
\hline
& \multicolumn{2}{c}{CUHK03-L} & \multicolumn{2}{|c|}{CUHK03-D} & \multicolumn{2}{c}{DukeMTMC} & \multicolumn{2}{|c|}{Market1501} \\
Method & R1 & mAP & R1 & mAP & R1 & mAP & R1 & mAP \\
\hline\hline
AOS~\cite{huang2018adversarially}&-&-&47.1&43.3&79.2&62.1&86.5&70.4\\
PCB~\cite{Sun_2018_ECCV}&-&-&63.7&57.5&83.3&69.2&93.8&81.6 \\
MGN~\cite{wang2018learning}&68.0&67.4&66.8&66.0&88.7&78.4&95.7&86.9 \\
Pyramid~\cite{zheng2018pyramidal}&78.9&76.9&78.9&74.8&89.0&79.0&95.7&88.2 \\
DG-Net~\cite{zheng2019joint}&-&-&65.6&61.1&86.6&74.8&94.8&86.0 \\
CAMA~\cite{yang2019towards}&70.1&66.5&66.6&64.2&85.8&72.9&94.7&84.5 \\
CASN~\cite{zheng2019re}&73.7&68.0&71.5&64.4&87.7&73.7&94.4&82.8 \\
BDB~\cite{dai2019batch}&73.6&71.7&72.8&69.3&86.8&72.1&94.2&84.3 \\
DSA~\cite{zhang2019densely}&78.9&75.2&78.2&73.1&86.2&74.3&95.7&87.6 \\
AA-Net~\cite{tay2019aanet}&-&-&-&-&87.7&74.3&93.9&83.4\\
IANet~\cite{ianet}&-&-&-&-&87.1&73.4&94.4&83.1 \\
MHN~\cite{chen2019mixed}&77.2&72.4&71.7&65.4&89.1&77.2&95.1&85.0 \\
${P}^2$-Net~\cite{guo2019beyond} &78.3&73.6&74.9&68.9&86.5&73.1&95.2&85.6 \\
OS-Net~\cite{zhou2019omni}&-&-&72.3&67.8&88.6&73.5&94.8&84.9 \\
ABD-Net~\cite{chen2019abd}&-&-&-&-&89.0&78.6&95.6&88.3 \\
FPR~\cite{he2019foreground}&-&-&76.1&72.3&88.6&78.4&95.4&86.6 \\
SCAL~\cite{chen2019self}&74.8&72.3&71.1&68.6&89.0&79.6&95.8&89.3 \\
CAR~\cite{zhou2019discriminative}&-&-&-&-&86.3&73.1&\textbf{96.1}&84.7 \\
\textbf{Ours}  &\textbf{82.4}&\textbf{79.6}&\textbf{80.2}&\textbf{77.1}&\textbf{90.0}&\textbf{80.7}&95.8&\textbf{89.6} \\
\hline
AOS*&-&-&54.6&56.1&84.1&78.2&88.7&83.3\\
AlignedRe-ID~\cite{zhang2017alignedreid}* &-&-&-&-&-&-&94.4&90.7 \\
MGN* &-&-&-&-&-&-&\textbf{96.6}&94.2 \\
TriNet~\cite{HermansBeyer2017Arxiv}* &-&-&-&-&-&-&86.6&81.1 \\
AA-Net* &-&-&-&-&90.4&86.9&95.1&92.4\\
SPT~\cite{luo2019spectral}* &-&-&-&-&88.3&83.3&93.5&90.6 \\
\textbf{Ours}*  &\textbf{87.3}&\textbf{89.1}&\textbf{86.3}&\textbf{87.5}&\textbf{91.8}&\textbf{90.6}&\textbf{96.5}&\textbf{95.1} \\
\hline
\end{tabular}
\end{center}
\caption{Comparison with other State-Of-the-Art methods. Re-ranking~\cite{zhong2017re, qin2011hello} is applied for method with ``*". Top performances on each dataset are in \textbf{bold}.}
\label{sota}
\end{table*}

Semantic mask labels are transformed from the Densepose~\cite{alp2018densepose} dataset. For Semantic Adversarial Branch, $U_{torso}$ contains \{head, upper arm, lower arm, chest\} and $L_{torso}$ includes \{upper leg, lower leg, foot\} respectively.

\subsection{Datasets and Protocols}
\textbf{Market1501}~\cite{zheng2015scalable} dataset contains 32,688 images of 1,501 identities captured by 6 cameras. The training set contains 751 identities with 12,936 images, the testing set contains 750 identities with 3,368 query images and 15,913 gallery images. 

\noindent\textbf{CUHK03-NP}~\cite{li2014deepreid} dataset contains two sub-sets, namely CUHK03-L(labelled set) and CUHK03-D(detected set) according to data generation method. The CUHK03-L contains 14,096 images and CUHK03-D contains 14,097 images. We adopt the new train / test split protocol proposed by~\cite{li2014deepreid}, which contains 767 identities for training and 700 identities for testing.

\noindent\textbf{DukeMTMC-reID}~\cite{zheng2017unlabeled, ristani2016MTMC} dataset is collected from 8 cameras. contains 16,522 training images of 702 identities, 222,8 query images and 17,661 gallery images in the testing set of the other 702 identities.

\noindent\textbf{Protocols.} We adopt two commonly used evaluation protocols in our experiments. Rank-1 identification rate (R1) and the mean Average Prevision (mAP).

\subsection{Comparison with other State-of-the-Arts}
We compare our method against 21 State-of-the-Art methods and present the results in Table~\ref{sota}. Our method achieves the best results on all the datasets except for Rank-1 on Market1501. Notably, our method outperforms other networks by a large margin on the small and challenging dataset CUHK03 with relatively heavy viewpoint limitation and occlusion. Furthermore, the MagnifierNet outperforms part-alignment related methods including BDB~\cite{dai2019batch} and ${P}^2$-Net~\cite{guo2019beyond}, which validates the effectiveness of our method to go beyond alignment and further improve Re-ID performance via semantic adversarial and fusion. In addition, the results also showcase the superiority of our approach over other methods with feature regularization such as AOS~\cite{huang2018adversarially}, as our method extracts attentive features accurately from each aligned semantic region and perform end-to-end adversarial training directly on the semantic level.
\subsection{Ablation Experiments}
We have carried out extensive ablation studies to validate the effectiveness of each module in MagnifierNet, which will be covered in the following sections.

\noindent\textbf{Improvement from each Network Component}   There are three important components in our framework: $M$ to impose alignment constraint, $SAB$ to encourage attentive semantic feature learning, and $SFB$ to filter and fuse learned representations. We add network components one by one onto the baseline which contains only the Global Branch. The performance improvements from each component are presented in Table~\ref{branch-abla}. It is shown that $SAB$ and $SFB$ are able to boost the performance both individually and simultaneously to a large margin. 
The proposed $SD$ Loss is able to attain further gain on all datasets, which validates its effectiveness across different domains. 

In addition, to further elaborate the benefits of each component qualitatively, we visualize the saliency maps of some randomly selected images using different settings in Figure~\ref{backbone-vis}. The baseline alone can only capture a coarse representation of the image. Both $SAB$ and $SFB$ are able to significantly improve model's activation on each semantic region, while $SAB$ tends to highlight every region and $SFB$ tends to selectively focus on certain semantic parts. When applying dual branches simultaneously, their individual advantages complement each other which yields a joint-representation that selectively highlights hidden details in each informative region.


 \begin{table*}[h]
\begin{center}
\begin{tabular}{|c c c c c|c c|c c|}
\hline
\multicolumn{5}{|c|}{Components}&
\multicolumn{2}{|c|}{CUHK03-L}& 
\multicolumn{2}{|c|}{DukeMTMC} \\
$G$&$M$&$SAB$&$SFB$&$SD$&R1&mAP&R1&mAP\\
\hline
\cmark&&&&&69.8&67.4&82.7&70.8\\
\cmark&\cmark&&&&71.3 (+1.5)&69.0 (+1.6)&83.9 (+1.2)&72.9 (+2.1)\\
\cmark&\cmark&\cmark&&&76.6 (+6.8)&74.6 (+7.2)&87.1 (+4.4)&76.7 (+5.9)\\
\cmark&\cmark&&\cmark&&77.4 (+7.6)&75.9 (+8.5)&86.7 (+4.0)&75.2 (+4.4)\\
\cmark&\cmark&\cmark&\cmark&&81.2 (+11.4)&78.5 (+11.1)&88.6 (+5.9)&79.6 (+8.8)\\
\cmark&\cmark&\cmark&\cmark&\cmark&82.4 (+12.6)&79.6 (+12.2)&90.0 (+7.3)&80.7 (+9.9)\\
\hline
\end{tabular}
\end{center}
\caption{Ablation study on the impact of different network components. We consider the Global Branch $G$ as the baseline performance. The improvement from each component over the baseline is indicated in the bracket.}
\label{branch-abla}
\end{table*}

\begin{figure*}[t!]
        \center{\includegraphics[width=1.0\textwidth]
        {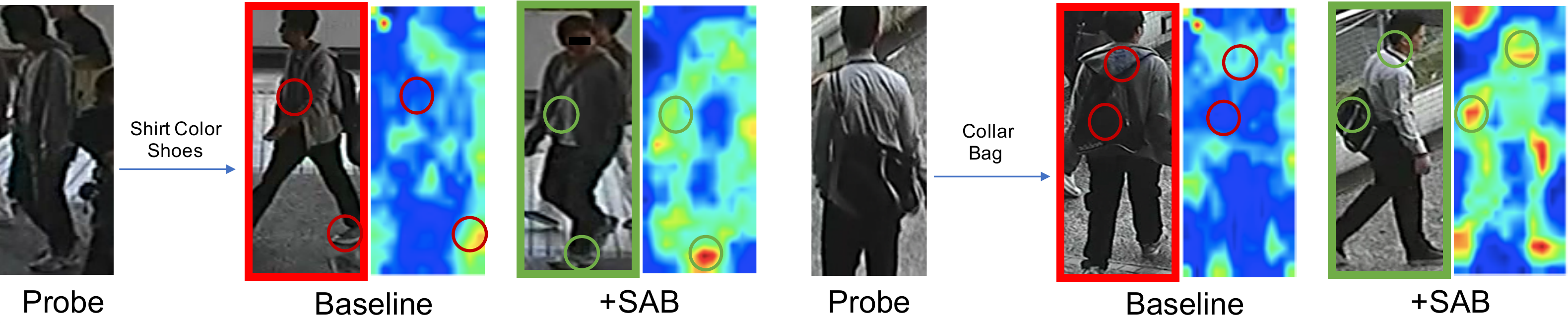}}
        \caption{\label{SRB-vis}Top 1 search results with Baseline and Baseline + $SAB$. The baseline fail to retrieve the correct image and capture distinctive features as shown in red, while $SAB$ manages to generate correct result and locate attentive features as shown in green.}
\end{figure*}
 
\noindent\textbf{Effect of Semantic Adversarial}   We visualize the top 1 querying results between the baseline and baseline + $SAB$ for two probe images in Figure~\ref{SRB-vis}. As shown in the feature activations, $SAB$ can significantly improve model's sensitivity on semantic details. For the first query on top, $SAB$ model magnifies the top 1 gallery image's upper body and foot region for comparison. For the second query at the bottom, $SAB$ model captures its top 1's collar and back region when retrieving. However, the baseline method produces incorrect results on these cases due to inability to highlight critical details as shown in its activation maps.

\begin{figure}[htbp] 
\centering
\begin{minipage}[t]{0.48\textwidth}
\centering
\includegraphics[width=6.2cm, height=0.23\textheight]{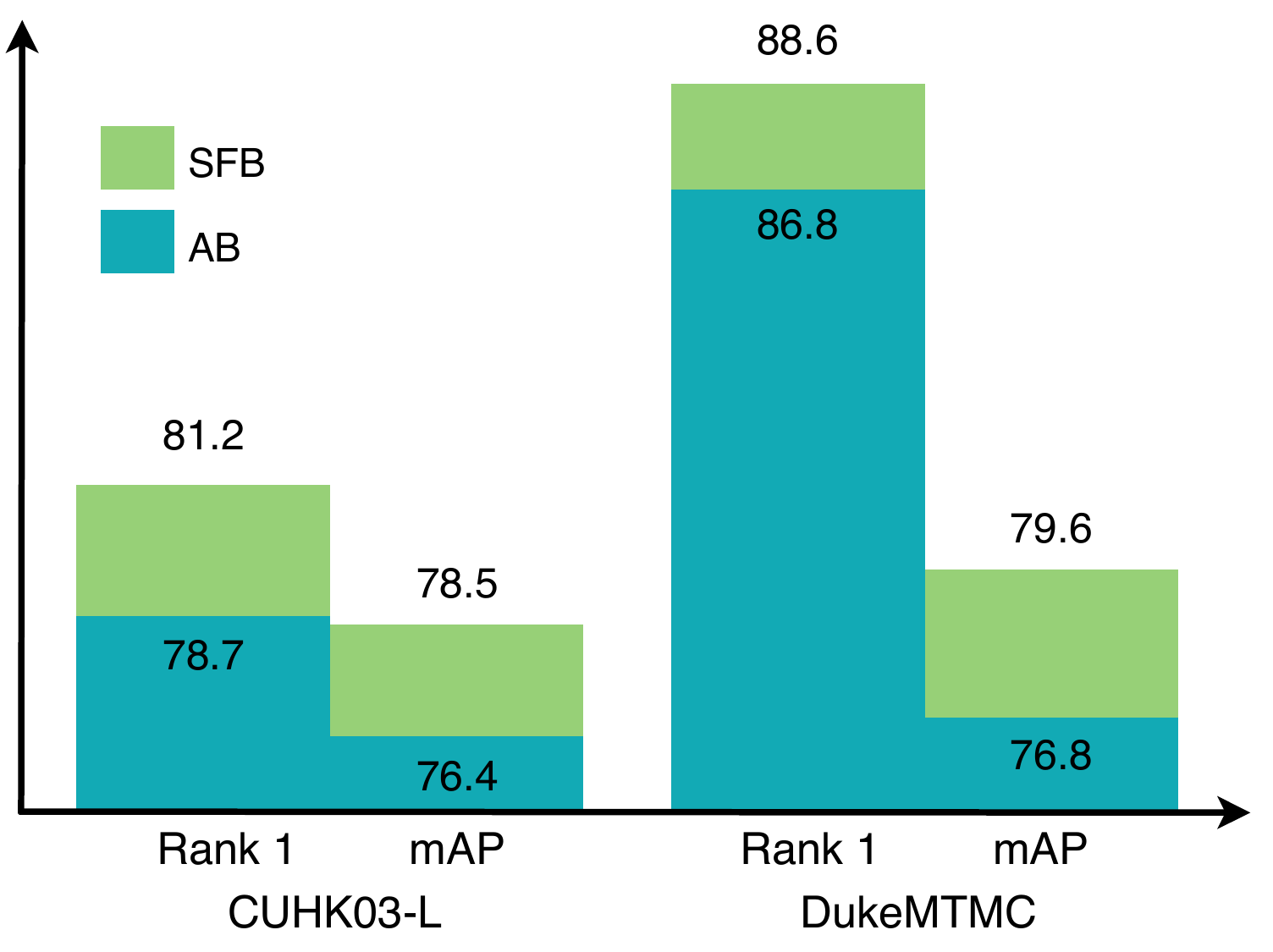}
\caption{\label{sfb-abla-quant}Comparison between $SFB$ and its Ablation Branch, where $SFB$ significantly improves both R1 and mAP.}
\end{minipage}
\begin{minipage}[t]{0.48\textwidth}
\centering
\includegraphics[width=6.5cm,height=0.25\textheight]{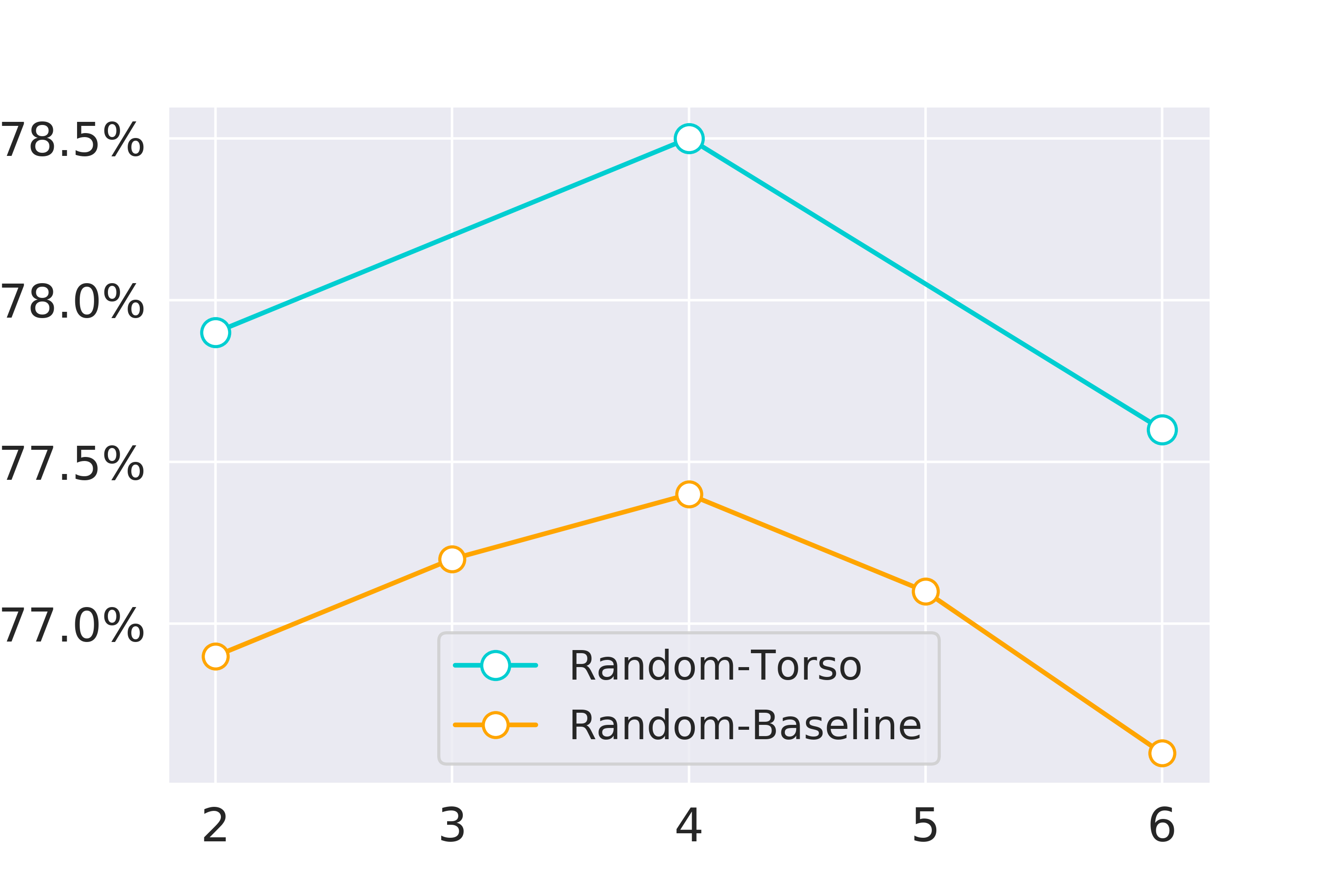}
\caption{\label{SAB-abla-torso}Effect of different number of remained semantic regions $\hat{K}$. Note that $\hat{K}$ is even for Random-torso as discussed in Section 3.2.}
\end{minipage}
\end{figure}

In addition, we also analyze different sampling strategies for the partial semantic representation. Specifically, we train MagnifierNet with different $\hat{K}$ in $SAB$ for CUHK03-L dataset. As shown in Figure~\ref{SAB-abla-torso}, the mAP of $SAB$ with \textbf{Random-torso} is consistently superior to that with \textbf{Random-baseline}.
We therefore demonstrate the effectiveness of our proposed sampling strategy,
while the best performance is both achieved when $\hat{K}$ is 4, which is the setting for $SAB$ in our experiments. The study is done without $SD$ Loss to highlight the impact of semantic adversarial itself.

\noindent\textbf{Effect of Semantic Fusion}   To validate the benefit of $SFB$, we replace it with an ``ablation branch" ($AB$) and compare their performances. To construct the $AB$, instead of feeding semantic features into GRU sequentially, we directly perform a $1\times1$ convolution to reduce their dimensions followed by concatenation to ensure $AB$ has the same output feature dimension as $SFB$. We train both models until converge without $SD$ Loss to validate the effectiveness of network structure alone. As shown in Figure~\ref{sfb-abla-quant}, the proposed semantic fusion method $SFB$ surpasses $AB$ on all metrics, which validates its positive influence on our framework.





\section{Conclusion}
In this paper, we propose a novel network MagnifierNet that improves ReID performance beyond pure alignment. The Semantic Adversarial Branch mines the fine-grained details in each semantic region by learning with limited semantic representation, the Semantic Fusion Branch selectively encodes semantic features by filtering out noises and focusing only on beneficial information. We further improve the model performance by introducing a novel Semantic Diversity Loss which promotes feature diversity among semantic regions. MagnifierNet achieves State-of-the-Art performance on three major datasets Market1501, DukeMTMC-reID, and CUHK03, which showcases the effectiveness of our method. 

\bibliography{egbib}
\end{document}